\documentclass{article} % For LaTeX2e
\usepackage{conference,times}

\usepackage{amsmath,amsfonts,bm}

\def\eqref#1{equation~\ref{#1}}
\def\1{\bm{1}}

\DeclareMathAlphabet{\mathsfit}{\encodingdefault}{\sfdefault}{m}{sl}
\SetMathAlphabet{\mathsfit}{bold}{\encodingdefault}{\sfdefault}{bx}{n}

\DeclareMathOperator*{\argmax}{arg\,max}

\usepackage{hyperref}
\usepackage{url}
\usepackage{booktabs}
\usepackage{tikz}
\usepackage{cleveref}
\usepackage{xcolor}
\usepackage{mdframed}
\usepackage{subcaption}
\usepackage{filecontents}
\usepackage{wrapfig}
\usepackage{soul}
\usepackage{pgfplots}

\usepackage{algpseudocode}
\usepackage{algorithm}
\usepackage{optidef}
\usepackage{cleveref}
\usepackage{amsmath}
\usepackage{amssymb}
\usepackage{graphicx}

\usepackage{tikz}
\newcommand{\myfancylabel}{\begin{tikzpicture}[every node/.style={rotate=45}]%
\node[fill,inner sep=0pt,minimum size=0.5ex] at (0ex,0.5ex) {};%
\node[fill,inner sep=0pt,minimum size=0.5ex] at (0ex,-0.5ex) {};%
\node[fill,inner sep=0pt,minimum size=0.5ex] at (0.5ex,0ex) {};%
\node[fill,inner sep=0pt,minimum size=0.5ex] at (-0.5ex,0ex) {};%
\end{tikzpicture}}

\pgfplotsset{compat=1.15}% current version is 1.15
\usepgfplotslibrary{fillbetween}

\definecolor{cs1}{HTML}{c7e9b4}
\definecolor{cs2}{HTML}{7fcdbb}
\definecolor{cs3}{HTML}{41b6c4}
\definecolor{cs4}{HTML}{1d91c0}
\definecolor{cs5}{HTML}{225ea8}
\definecolor{cs6}{HTML}{253494}
\definecolor{cs7}{HTML}{081d58}

\definecolor{cd2}{HTML}{addd8e}
\definecolor{cd3}{HTML}{78c679}
\definecolor{cd4}{HTML}{41ab5d}
\definecolor{cd5}{HTML}{238443}
\definecolor{cd6}{HTML}{006837}
\definecolor{cd7}{HTML}{004529}
\definecolor{ForestGray}{HTML}{004529}

\definecolor{codegreen}{rgb}{0,0.6,0}
\definecolor{codegray}{rgb}{0.5,0.5,0.5}
\definecolor{codepurple}{rgb}{0.58,0,0.82}
\definecolor{backcolour}{rgb}{0.95,0.95,0.92}
\definecolor{jcred}{HTML}{e31a1c}
\definecolor{jcgreen}{HTML}{33a02c}
\definecolor{jcblue}{HTML}{1f78b4}
\definecolor{jcorange}{HTML}{ff7f00}
\definecolor{jcpurple}{HTML}{6a3d9a}
\definecolor{jclightred}{HTML}{fb8072}
\definecolor{jclightgreen}{HTML}{b3de69}
\definecolor{jclightblue}{HTML}{80b1d3}
\definecolor{jclightorange}{HTML}{fdb462}
\definecolor{jclightpurple}{HTML}{bebada}
\definecolor{jcredl}{HTML}{fb8072}
\definecolor{jcgreenl}{HTML}{b3de69}
\definecolor{jcbluel}{HTML}{80b1d3}
\definecolor{jcorangel}{HTML}{fdb462}
\definecolor{jcpurplel}{HTML}{bebada}
\definecolor{jcyellow}{HTML}{fdb462}

\colorlet{definitioncolour}{ForestGray!10!white}

\mdfdefinestyle{fancyenv}{
    skipabove=0.8\baselineskip,
    skipbelow=0.\baselineskip,
    innertopmargin=8pt,
    innerbottommargin=8pt,
    linewidth=2pt,
    topline=true,
    frametitleaboveskip=\dimexpr-\ht\strutbox\relax,
    nobreak=true,
}

\newcounter{definition}

\title{ARIES: Autonomous Reasoning with LLMs on Interactive Thought Graph Environments}

\author{
Pedro Gimenes$^1$~\thanks{Corresponding author.}, Zeyu Cao$^2$, Jeffrey Wong$^1$, Yiren Zhao$^1$ \\
$^1$Department of Electrical \& Electronic Engineering,
Imperial College London \\
$^2$Department of Computer Science and Technology, University of Cambridge \\
\texttt{\{pedro.gimenes19, tsz.wong20, a.zhao\}@ic.ac.uk} \\
\texttt{zeyu.cao@cl.cam.ac.uk} \\
}

\iclrfinalcopy
\begin{document}

\maketitle

\begin{abstract}
Recent research has shown that LLM performance on reasoning tasks can be enhanced by scaling test-time compute. One promising approach, particularly with decomposable problems, involves arranging intermediate solutions as a graph on which transformations are performed to explore the solution space. However, prior works rely on pre-determined, task-specific transformation schedules which are subject to a set of searched hyperparameters. In this work, we view thought graph transformations as actions in a Markov decision process, and implement policy agents to drive effective action policies for the underlying reasoning LLM agent. In particular, we investigate the ability for another LLM to act as a policy agent on thought graph environments and introduce ARIES, a multi-agent architecture for reasoning with LLMs. In ARIES, reasoning LLM agents solve decomposed subproblems, while policy LLM agents maintain visibility of the thought graph states, and dynamically adapt the problem-solving strategy. Through extensive experiments, we observe that using off-the-shelf LLMs as policy agents with no supervised fine-tuning (SFT) can yield up to $29\%$ higher accuracy on HumanEval relative to static transformation schedules, as well as reducing inference costs by $35\%$ and avoid any search requirements. We also conduct a thorough analysis of observed failure modes, highlighting that limitations on LLM sizes and the depth of problem decomposition can be seen as challenges to scaling LLM-guided reasoning.
\end{abstract}

\section{Introduction}

% Why to do test time compute scaling
Prior works have shown that Large Language Models (LLMs) are subject to the emergence of abilities as their parameter count grows \citep{emergent_abilities}, which spurred significant interest in training increasingly larger models. However, recent work showed that under a fixed compute budget for training and inference, LLM performance on reasoning tasks can be enhanced by allocating a higher proportion of compute to inference rather than training \citep{scaling}. This shift towards inference-time compute scaling can be intuitively understood through the Dual Process Theory, which postulates the existence of two distinct modes of reasoning in humans - (1) a fast, intuitive mode and (2) a slow, deliberate mode \citep{dual}. While the autoregressive decoding procedure of LLMs resembles System 1, prior works used LLMs in System 2 reasoning by inducing models to thoroughly explore a problem, such as using chain of thoughts, ahead of providing a solution to the user query \citep{cot}.

% , or by viewing task accuracy as a reward signal on which to train models via Reinforcement Learning \citep{r1}

% Contrasting methods: finetuning on reasoning traces vs structure-enhanced
System 2 reasoning can be induced in LLMs by querying models fine-tuned on extensive reasoning traces \citep{s1}. While such single-query approaches have been shown effective in improving the quality of complex sequential logic, an alternative approach involves \emph{performing multiple queries with the same LLM} and arranging intermediate solutions (or ``thoughts") in a specified topology, i.e. topological reasoning \citep{demistifying}. This approach yields benefits in problems where intermediate solutions can be reliably scored through a Process Reward Model (PRM) \citep{scaling} or using real feedback from external environments \citep{tot}. Additionally, a graph formulation has shown promising results in problems displaying the property of decomposability into subproblems that can be solved independently then aggregated through a sequence of graph transformations \citep{got}. In this work, we focus on \textit{problems with the decomposability property} and in environments where external feedback is viable and useful, such as using LLMs to solve coding problems.

% work from here
% Figure X shows an example of this approach. (coding task, write skeleton, then implement each subfunction)

Despite the benefits of topological reasoning, prior works rely on pre-determined traversal strategies parametrized by a discrete set of hyperparameters. This approach lacks generality, as these parameters must be tuned manually or through extensive Bayesian search to achieve high query efficiency, due to the varying characteristics of each task. With this limitation in mind, we hypothesize that the generalization of artificial problem-solving towards (or beyond) human-like abilities in arbitrary domains requires a mechanism for autonomous traversal of a solution space, falling outside the constrained scope of static schedules shown in Tree-of-Thoughts \citep{tot} and Graph-of-Thoughts \citep{got}. 

To this end, we propose viewing thought graphs as an interactive environment where a sequence of graph transformations is seen as actions in a Markov Decision Process (MDP). Considering this state-action formulation, an effective action policy should explore the solution space to yield a solution while learning from external feedback. Such a mechanism would present a step towards general intelligent agents capable of leveraging existing world knowledge while adapting to out-of-distribution tasks.

Motivated by recent improvements in LLM planning and reasoning \citep{cot, react}, we aim to investigate whether existing LLMs have the capability to act as autonomous reasoning agents by formulating thought graphs as interactive environments. We propose the use of LLM policy agents (i.e. LLM-based action planners) to autonomously execute a set of transformations, including thought proposal, evaluation, aggregation and refinement. 
As such, we consider the following research questions: \textbf{(1)}~Can LLMs act as policy agents and effectively utilize feedback from thought graph environments to dynamically tune their exploration strategies? \textbf{(2)}~Can this approach match the performance of static transformation schedules extensively optimized for a given task? And finally, \textbf{(3)}~What are the failure modes of using existing LLMs as policy agents in guiding thought graph exploration (i.e.~factors affecting the ability to produce coherent exploration plans)?

% The tree exploration in ToT can be tuned by parameters $(w, d, n)$ where $w$ is the tree width (i.e. number of generated proposals at each level), $d$ is the depth (i.e. number of layers) and $n$ is the number of proposals kept at each level.
% \az{Aaron's marker to come back to this tomorrow morning.}

% Figure \ref{fig:introduction} presents an example of our findings regarding question \textbf{(1)}, by contrasting an extracted trace of our approach against static tree search and graph transformation schedules. In tree search, each node is an attempted solution, with the best-scoring nodes kept and refined at each tree level \citep{tot}. The static graph transformation schedule involves decomposing the starting problem into smaller subproblems, solving them individually then merging the intermediate results \citep{got}. 

% In our approach, we leverage the planning abilities of LLMs, using them as policy agents to choose transformation actions given the state of a thought graph. 

% We observe the policy agent displays \textbf{dynamic self-correction through adaptive decomposition}, meaning the depth of decomposition and query utilization are dynamically tuned according to the complexity of the task.

\begin{figure*}
    \centering
    \includegraphics[width=\linewidth]{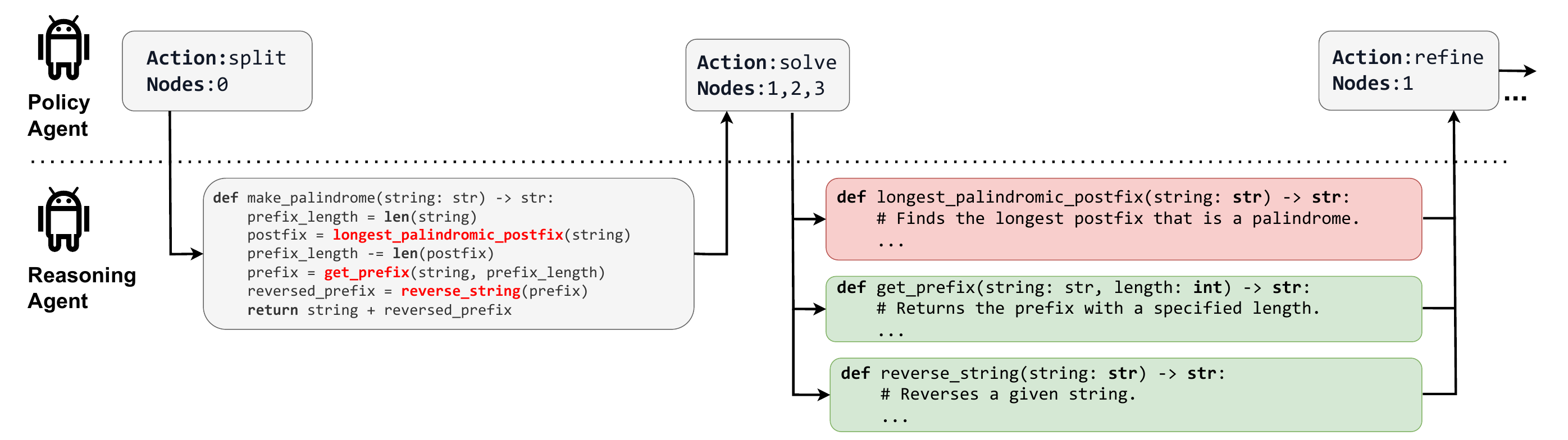}
    \caption{ARIES workflow in answering the HumanEval prompt: "Find the shortest palindrome that begins with a supplied string". The policy agent selects an action based on the thought graph state, which is executed by the reasoning agent. First, the split action generates a skeleton implementation calling yet-to-implement subfunctions, decomposing the problem. Then, the agent is instructed to generate a solution for each subfunction. Since one of the solutions doesn't pass its testcases, the reasoning agent is instructed to refine it based on execution feedback.}
    \label{fig:introduction}
\end{figure*}

We investigate the aforementioned questions by implementing ARIES, a multi-agent framework for solving reasoning problems formulated as thought graphs. Figure \ref{fig:introduction} provides a summary of our approach - in each iteration, the policy agent monitors the thought graph state and samples from the action space to choose a graph transformation. The reasoning agent then performs these transformations and updates the thought graph state. In summary, our contributions are as follows.

\begin{itemize}
    \item We introduce ARIES, a novel formulation to autonomous topological reasoning, making the whole reasoning task LLM-guided. We frame the topological reasoning task as a collaboration between two agents within a topological thought graph. The LLM policy agent assesses states and determines the actions, while the LLM reasoning agent carries out these actions, executing transformations on the thought graph.
    
    \item We show that LLMs exhibit planning capacity and can serve effectively as policy agents on topological reasoning tasks, thus eliminating the requirement for predefined, task-specific scheduling of the reasoning agents, as seen in Tree-of-Thoughts (ToT) and Graph-of-Thoughts (GoT). Additionally, we identify and discuss the limitations and failure modes of their planning abilities.

    \item We perform carefully controlled experiments against a number of benchmarks, showing that LLM-guided thought graph exploration can lead to up to $29\%$ higher accuracy at $35\%$ lower inference cost, as well as obviating any Bayesian search cost.
\end{itemize}

\section{Related Work} \label{section:related-work}

% todo: explanation of SER as method for test time scaling

\subsection{Topological Reasoning}

\citet{cot} pioneered the elicitation of step-by-step logical reasoning, with subsequent work by \citet{cot_sc} demonstrating improved performance through the sampling and arbitration along multiple reasoning sequences. \citet{tot} formulate concurrent exploration of multiple reasoning paths by scoring reasoning steps, leveraging tree search algorithms (ToT). Finally, \citet{got} generalize problem space exploration by formulating thoughts as a graph, enabling the use of arbitrary transformations such as node refinement and aggregation (GoT).

Several works have explored methods of improving the query efficiency of topological reasoning, which suffers from high computational demand due to iterative LLM prompting \citep{tomt, aot, xot}. Despite improvements, few works have targeted the generality of this approach by exploring dynamic transformations. While \citep{tot} leverage standard tree search algorithms, \citep{tot2} hypothesize that tree search can be enhanced through trained policy networks to guide node backtracking. However, this idea is not explored fully and their evaluation is focused on heuristics-based rules. As such, our work presents the first effort towards generalized topological reasoning through autonomous thought graph exploration.

% \citep{tomt} proposed combining one-shot generation (i.e.~LLM decoding without iterative prompting) with ToT search in visual reasoning benchmarks. \citep{aot} compresses thought graph exploration traces as in-context examples, prompting LLMs to reproduce the exploration in fewer queries, which shows promising results despite not being suitable for interactive environments. Finally, \citep{xot} combines pre-trained reinforcement learning with Monte-Carlo search on thought graphs to generate solution paths with minimal LLM intervention, although this requires extensive training and lacks task generality.

\subsection{LLMs as Action Policy Agents}

Significant research has focused on leveraging LLMs for guiding action policies, such as in tasks requiring coordination of heterogeneous model ensembles \citep{hugginggpt}. LLMs have also been deployed as action planners in interactive environments where feedback is provided to the action scheduler, such as solving computer tasks \citep{llm_computer_tasks} and online shopping \citep{react}. However, some works have outlined the instability in obtaining action plans over long-range horizons, where LLMs have been shown to repeatedly generate invalid action plans~\citep{xie2023translatingnaturallanguageplanning}. This limitation has been tackled by works such as \citep{reflexion}, which propose an episodic memory buffer of previous trials. However, to our knowledge, no prior work has investigated leveraging LLM planning abilities in the context of topological reasoning.

\section{Topological Reasoning with Large Language Models} \label{section:topological-reasoning}

We consider a reasoning problem to be stated in language as an ordered tuple of tokens $p = (t_1, \dots, t_m)$, where each token $t \in \mathcal{V}$ belongs to a vocabulary space $\mathbb{V}$. We define a thought $\tau = (t_1, \dots, t_j)$ as a sequence of tokens sampled autoregressively from an LLM parametrized by $\theta$, i.e.  $t_i \sim P(t_i \mid t_1, \dots, t_{i-1}; \theta)$. This consists of a language representation of an intermediate step towards the solution to the problem.

A thought sequence can be represented as an ordered tuple of thoughts $S = (\tau^1, \tau^2, \dots, \tau^k)$ of length $k$, such that the final thought $\tau^k$ represents a candidate solution to the problem $p$. A thought tree $T_\tau$ can be represented as $(\mathcal{V}, \mathcal{E})$, where $\mathcal{V}$ is a set of thought nodes and $\mathcal{E}$ is a set of edges connecting them. The tree can be parametrized with a depth of $d$ and a width of $w$, denoting the number of nodes per level. Additionally, each thought $\tau^{ij}$ ($j$-th thought at depth $i$) has a value $\lambda (\tau^{ij})$ such that nodes with higher values yield valid solutions to the problem with higher probability. Hence, tree-based thought exploration involves finding a path $\mathcal{P} \subset \mathcal{V}$ that maximizes the cumulative value of thoughts, as follows. 
\begin{equation} \label{eq:tot}
    P^* = \argmax_{\mathcal{P}} \underset{\tau \in \mathcal{P}}{\sum} \lambda(\tau)
\end{equation}

A thought graph $G_\tau$ can also be represented via the tuple $(\mathcal{V}, \mathcal{E})$, with no imposed restriction on the arrangement of thoughts and edges. Thought graph exploration can be regarded as a sequence of $m$ graph transformations as follows, where each $\phi_i: G_\tau^i \rightarrow G_\tau^{i+1}$ modifies the set of nodes and edges. The full set of considered transformations and their formulations are shown in Table~\ref{table:transformations}.
\begin{equation} \label{eq:transformations}
    G^*_\tau = \phi_m ( \dots ( \phi_1 ( \phi_0 (G^0_\tau))))
\end{equation}

Table \ref{table:transformations-simple} summarizes the thought graph transformations we consider in the rest of this work. $\phi_{dec}$ decomposes a reasoning problem into subproblems to be solved individually, creating new nodes in the thought graph. $\phi_{sol}$ generates a candidate solution to a subproblem. $\phi_{ref}$ considers an incorrect subproblem solution, utilizing further LLM queries to refine it. $\phi_{red}$ removes nodes in the graph according to their values. Finally, $\phi_{agg}$ performs node merging to aggregate subproblem solutions into a coherent solution to the original problem.

\begin{table}[!t]
    \centering
    \caption{Thought graph transformations used to solve reasoning problems using a divide-and-conquer strategy. See Appendix \ref{appendix:thought_graph_transformations} for their complete definitions.}
    \label{table:transformations-simple}
    \vspace{0.5em}
    
    \begin{tabular}{lc}
    \toprule
    \textbf{Transformation} & \textbf{Symbol} \\ 
    \midrule
    
    {Decompose} & $\phi_{dec}$ \\
    
    {Solve} & $\phi_{sol}$ \\

    {Refine} & $\phi_{ref}$  \\
    %  \midrule
    
    {Reduce} & $\phi_{red}$  \\
    % \midrule
    
    {Aggregate} & $\phi_{agg}$ \\
    \bottomrule
    \end{tabular}
\end{table}

\textbf{Static Transformation Schedules:} A static transformation schedule can be parametrized by the tuple $(R_{ed}, R_{ef}, S^m, A^m, R_{ef}^m)$. $S^m, A^m, R_{ref}^m$ represents the multiplicity (i.e. number of attempts) of the solve, aggregate and refine transformations, respectively. $R_{ed}, R_{ef} \in \{0, 1\}$ indicate whether the $\phi_{red}$ and $\phi_{ref}$ transformations are applied after aggregation. 

\begin{algorithm}[!t]
    \caption{Static Thought Graph Transformation Schedule}
    \label{algo:got-schedule}
    \begin{algorithmic}
        \Require~Starting graph $G_\tau^0$, allow reduce $R_{ed}$, allow refine $R_{ef}$
        \Require~Solve multiplicity $S^m$, aggregate multiplicity $A^m$, and refine multiplicity $R_{ef}^m$

        % Decompose
        % \Comment{Decompose into subproblems}
        
        \State $G_\tau^{dec} \gets \phi_{dec} (G_\tau^0, 1, \{ 0\}))$  

        % Solve
        % \Comment{Solve subproblems}
        
        \State $G_\tau^{sol} \gets \phi_{sol} (G_\tau^{dec}, S^m, \Delta(G_\tau^{dec}, G_\tau^0))$ 

        % Aggregate
        % \Comment{Aggregate subproblem solutions}
        
        \State $G_\tau^{agg} \gets \phi_{agg} (G_\tau^{sol}, A^m, \Delta (G_\tau^{sol}, G_\tau^{dec}))$ 
        
        % Reduce
        % \Comment{Reduce aggregation attempts}
        
        \If{$R_{ed}$}
            \State $G_\tau^{red} \gets \phi_{red} (G_\tau^{agg}, 1, \Delta(G_\tau^{agg}, G_\tau^{sol}))$ 
        \Else
            \State $G_\tau^{red} \gets G_\tau^{agg}$
        \EndIf
        
        % Refine
        % \Comment{Refine aggregation attempts}
        
        \If{$R_{ef}$}
            \State $G_\tau^{ref} \gets \phi_{ref} (G_\tau^{red}, R_{ef}^m, \Delta(G_\tau^{red}, G_\tau^{agg}))$ 
            \State $G_\tau^* \gets \phi_{red} (G_\tau^{ref}, 1, \Delta(G_\tau^{ref}, G_\tau^{red}))$ 
        \Else
            \State $G_\tau^* \gets G_\tau^{red}$
        \EndIf
        
        \State {\bf Return:}~$G_\tau^*$
    \end{algorithmic}

\end{algorithm}

In Algorithm \ref{algo:got-schedule}, each transformation is defined as $\phi (G_\tau, m, S)$, where $G_\tau = (V, E)$ is a thought graph, $S \subset V$ is a subset of nodes and $m$ is the multiplicity (number of attempts). Additionally, the function $\Delta(G_\tau^a, G_\tau^b)$ outputs all nodes present in the first graph $G_\tau^a = (\mathcal{V}_a, \mathcal{E}_a)$ but not in the second $G_\tau^b = (\mathcal{V}_b, \mathcal{E}_b)$, defined formally as follows.
\begin{equation}
    \Delta(G_\tau^a, G_\tau^b) = \{v | v \in \mathcal{V}_1 \And v \notin \mathcal{V}_2 \}    
\end{equation}

Algorithm \ref{algo:got-schedule} represents a standard divide-and-conquer strategy. The $\phi_{dec}$ transformation decomposes the starting problem into $B$ subproblems, which are solved individually ($\phi_{sol}$). The aggregation of the subproblem solutions is attempted $A^m$ times, as the $\phi_{agg}$ transformation has a non-zero probability of failure. If $R_{ed} = 1$, a single aggregation attempt is kept, while others are removed from the graph. If $R_{ef} = 1$, the remaining aggregation attempts are then refined wth $\phi_{ref}$, and the highest-scoring attempt is kept as the final solution.
        
\section{Thought Graph Exploration as a Markov Decision Process} \label{section:autonomous-graph-exploration}

\begin{figure*}
    \centering
    \includegraphics[width=\linewidth]{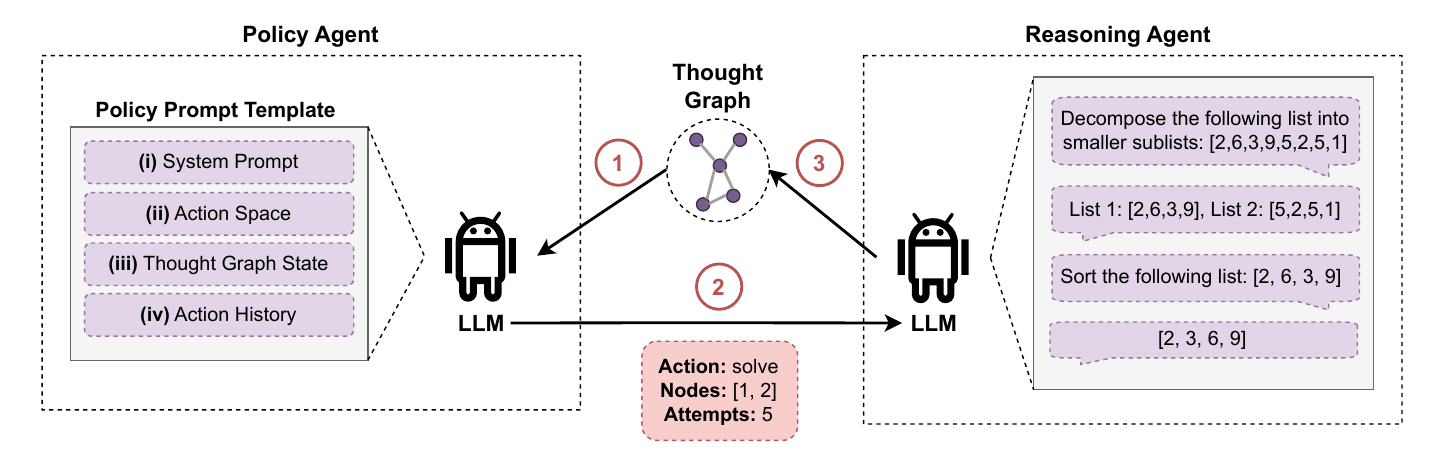}
    \caption{Multi-agent framework for reasoning over thought graphs. First, (1) the policy agent an action and subset of nodes given a prompt including (i-ii) general instructions and (iii-iv) an overview of the exploration state. The sample is then (2) passed to the reasoning agent, which finally (3) updates the thought graph state.}
    \label{fig:framework}
\end{figure*} 

Beyond the fixed schedule shown in Algorithm \ref{algo:got-schedule}, the transformation of a thought graph can be generalized as a Markov decision process $(\mathcal{S}, \mathcal{A}, \mathcal{P}_a)$:
\begin{itemize}
    \item \textbf{State} $s_t \in \mathcal{S}$: represents an arrangement of nodes and edges in the thought graph, with the associated value of each node, i.e. $s_t = (\mathcal{V}, \mathcal{E}, \{\lambda(v) | v \in \mathcal{V} \} )$.
    
    \item \textbf{Action} $a \in \mathcal{A}$: indicates which transformation to perform on the thought graph, and which nodes to perform it on, i.e. $\mathcal{A} = \{(\mathcal{V}_s, \phi) \mid \mathcal{V}_s \subset \mathcal{V}, \phi \in \Omega\}$, where $\Omega$ is the set of transformations (Table \ref{table:transformations}).
    
    \item \textbf{Transition probability} $\mathcal{P}_a (s, s')$: represents the probability that an action $a$ applied at state $s$ yields the expected new state $s'$.

    % \item \textbf{Transition probability} $\mathcal{P}_a (s_t, s_t')$: represents the probability that an action $a$ applied at state $s_t$ yields the expected new state $s_t'$.
        
    % \item \textbf{Action reward} $\mathcal{R}_a (s, s')$: represents an immediate reward value for the $s \rightarrow s'$ state transition, triggered by action $a$.
\end{itemize}

% Given a starting thought graph $G_\tau = (\{v\}, \emptyset)$ corresponding to starting state $s_0 = (\{v\}, \emptyset, \{\lambda(v)\})$, where $v$ represents the starting problem, the action $\phi_{sol}$ yields a solution state $s'$ with probability $P(s' \mid s_0, \phi_{sol})$. A solution state is any state with at least one node $v'$ that has $\lambda(v') = 1$, e.g. $s' = (\{v, v'\}, \{(v, v')\}, \{\lambda(v), 1\})$. 

The optimal transformation sequence $\Phi$ is then defined as the sequence of actions that maximize the conditional probability of reaching a solution state $s^+$, i.e. $\Phi = (\phi_0, \dots, \phi_n)$ that solves the following optimization problem.
\begin{maxi*}|s|
{\Phi}{P(s^+ \mid s^0, \Phi)}
{}{}
\addConstraint{|\Phi|}{< \epsilon}
\end{maxi*}

We bound the number of queries by the constant $\epsilon$, as in the limit $|\Phi| \rightarrow \infty$, $P(s^+ | s^0, \Phi) \rightarrow 1$. 
% \az{I think your $\Phi$ is a sequence of multiple $\Phi_{xx}$ defined in table 1, but this does not come out super clear here. I know you have a sentence explaining it, but might be worth to have a conrete math def.}

\subsection{Multi-Agent Reasoning}

In this work, we hypothesize that LLMs can approximate a solution to the stated optimization problem by acting as policy agents. We develop an interactive framework consisting of a policy agent and a reasoning agent, as shown in Figure \ref{fig:framework}. In each iteration, (1) the policy agent selects an action from the action space, (i.e. the transformations in Table \ref{table:transformations}). The policy agent then (2) directs the reasoning agent to perform the selected action. Finally, (3) the reasoning agent updates the thought graph. The process is repeated until a solution is found or a maximum number of iterations is reached.

The policy agent is invoked using the prompt template shown in Figure \ref{fig:framework}. (i) The system prompt outlines the problem setting, input format and expected behaviour from the policy agent. (ii) A task-specific list of actions, describing the preconditions and effects of each transformation, provides a semantic understanding of the action space. (iii) The current state of the graph is provided in a textual format, enumerating all nodes and edges. Finally, (iv) the action history in the current trial is included, promoting continuity in the strategies outlined in previous steps.

\subsection{In-Context Action Selection}

Prior work has shown that reasoning abilities of LLMs are enhanced when prompted to output a verbose sequence of steps before the solution \citep{cot, cot_sc}. This mechanism can be seen as enabling in-context task learning from some extracted innate world knowledge. Hence, our policy agent is instructed to generate a detailed analysis on the state of the thought graph and exploration history before sampling the action space. The analysis includes the following:
\begin{enumerate}
    \item Describe the action history and how each action relates to an exploration strategy.
    
    \item Describe the thought graph state, and how each node corresponds to previous actions.
    
    \item Discuss the outlined strategy, stating whether it is successful, unsucessful, or pending.
    
    \item Outline a number of options for the next action, detailing the expected outcome of each.
\end{enumerate}

% \subsection{Strategy Feedback Sampling}

% A critical use case for topological reasoning with LLMs involves solving the same problem repeatedly with different data. In this scenario, the model can learn from past experiences to improve its reasoning performance. We consider both positive and negative feedback, i.e. examples are included from both successful and unsuccessful trials. As such, efficient strategies are rewarded, while inefficient strategies are outlined to be avoided. Considering the context length constraint in LLMs, trial feedback requires an efficient sampling method to select a limited number of examples that maximize the information gain. This can be achieved by separating trials into buckets and randomly sampling a fixed number from each bucket, providing a comprehensive and diverse set of examples for tuning the exploration strategy.

\subsection{Policy Agent Ensembles}

Given the stochastic nature of token prediction in LLMs, we observe high variability in the chosen action over several invocations of a policy agent under the same thought graph state. Given the preconditions and effects of each action are represented via text rather than any rigorous formulation, actions selected by the policy agent can display flawed understanding of the problem constraints, leading to ineffective exploration of the thought graph. To overcome this limitation, we democratize action selection over an ensemble of agents, meaning a parametrizable number of LLM queries are performed concurrently at every iteration. The selected action is takes as the most frequent proposal among the ensemble. See \Cref{section:ablation} for ablation studies on the impact of policy agent ensemble size on reasoning performance.

% \az{someone may ask for an ablation on this: ensemble vs. single, this is something you may want to add in appendix.}

\section{Experiments} \label{section:experiments}

Through a range of controlled experiments, we evaluate the performance of LLM policy agents on interactive thought graphs. In \Cref{section:benchmarks} and \Cref{section:baselines}, we define the benchmarks and baselines. We present the core results across each benchmark task in \Cref{section:main-results}. We profile the transition probabilities of each thought graph transformation across tasks in \Cref{section:transition-probabilities}. In \Cref{section:failure-modes}, we provide empirical results demonstrating two main failure modes of LLMs as policy agents, namely model size and decomposition depth.

\textbf{Experimental Setup:} We evaluate Llama-3.1-70B and Llama-3.1-405B as policy and reasoning agents, hosted with SGLang at a temperature of~1. Llama-3.1-70B was hosted with $8\times$ A6000 GPUs. Llama-3.1-405B was hosted using $16\times$ H100 GPUs distributed over 4 nodes. The total cost was approximately 3k GPU hours.

% Claude-3.5-Haiku and GPT-4o were accessed using the OpenRouter platform, at a total cost of 2000USD.

\subsection{Benchmarks} \label{section:benchmarks}

We run our main evaluation on HumanEval, a widely used benchmark for assessing the functional correctness of code generation models through a set of Python programming problems with corresponding test cases \citep{human_eval}. Additionally, we consider two popular tasks for topological reasoning with LLMs, list sorting and set intersection. Despite their simplicity, prior works have shown that these tasks are extremely challenging for LLMs with direct prompting \citep{got}, benefitting from a divide-and-conquer strategy (i.e. decomposition, solving subproblems and merging). We evaluate these at various levels of difficulty (quantified by the size of the lists and sets), resulting in six benchmarks: sorting32/64/128 and set-intersection32/64/128.

For HumanEval, we report the task accuracy, while for list sorting and set intersection we report error function value $\mathcal{E}$. Details on the definition for the error function for each task can be found in Appendix \ref{section:benchmarks}. Additionally, we report both the search $C_s$ and inference cost $C_i$. We measure cost by the number of queries since we observe a low standard deviation in the number of generated tokens across all LLM queries during our experiments.

% to do: add appendix with statistics on number of generated tokens
\begin{table}[th]
    \centering
    \caption{Task accuracy ($\uparrow$), search and inference costs ($\downarrow$) on Human Eval. Cost is measured as the number of LLM queries. IO refers to direct prompting. Llama-405b was used for the reasoning and policy agents. }
        \begin{tabular}{c|ccc}
        \toprule
        & Accuracy & Search & Inference
        \\
        \textbf{Method}  & $[\%]$ & Cost $(C_s)$ & Cost $(C_i)$
        \\
        \midrule
        IO
        & 77.4
        & 0
        & 1 \\
        $\text{GoT}_{25\%}$
        & 66.3
        & 1160 
        & 34.8 \\
        $\text{GoT}_{50\%}$
        & 67.5
        & 2368 
        & 24.3 \\
        $\text{GoT}_{100\%}$
        & 60.1
        & 4742 
        & 8.17 \\
        ARIES
        &\textbf{ 89.0}
        & \textbf{0} 
        & \textbf{5.3} \\
        \bottomrule
        \end{tabular}
        \label{table:positive_results}

\end{table}

\subsection{Baselines} \label{section:baselines}

We use static transformation schedules as the baseline, following \citep{got}. As previously noted, static schedules require extensive, task-dependent hyperparameter tuning. For each individual task, we carefully tune the hyperparameters using Bayesian optimization resulting in three variants: $\text{GoT}_{25\%}$, $\text{GoT}_{50\%}$ and $\text{GoT}_{100\%}$. Here, the percentage corresponds to the number of trials spent until the hyperparameter search converges. As such, we compare against baselines with several search compute budgets. See~\Cref{section:search} for details on the full search methodology. We also consider an Direct IO (Input-Output) baseline, i.e. reasoning via direct LLM prompting.

\subsection{Evaluation} \label{section:main-results}

Replacing static transformation schedules with LLM policy agents offers generalization to arbitrary tasks at no tuning cost. However, performance is constrained by the LLM's planning capabilities. As such, we evaluate ARIES against the aforementioned benchmarks, demonstrating its advantages and identifying potential failure modes. We set the policy agent ensemble size to 5 in all experiments, as explained in Section \ref{section:ablation}.

\subsubsection{HumanEval}

Our key findings for autonomous policy agents in the context of a coding task are shown in  \Cref{table:positive_results}. It can be seen that by formulating this code generation task as a Markov decision process with an off-the-shelf LLM policy agent, we achieve up to $28.9\%$ higher accuracy than the most query-efficient static schedule baseline. We also observe that as further trials are expended in the GoT baseline search, the query efficiency is increased, i.e. hyperparameter configurations are found that achieve similar performance levels at lower query counts. Nevertheless, we achieve $54\%$ lower inference cost on average compared to even the most optimized GoT baseline, and also avoids any search time requirement.

% The benefits of the LLM policy agent can be particularly observed against the baselines with low search compute budget. In sorting32, ARIES leads to a $3.3\times$ improvement relative to $\text{GoT}_{\text{25\%}}$. In set-intersection32, we observe a $2.33\times$ improvement relative to $\text{GoT}_{\text{25\%}}$. ARIES surpasses even the most robust baseline ($\text{GoT}_{100\%}$) by $1.3\times$ for sorting32 and $3\times$ for set-intersection32, although both these baselines require substantial search cost. The absence of any search costs associated with ARIES makes it exceptionally appealing, serving as evidence that LLMs can function effectively as policy agents in reasoning tasks, and facilitating automated reasoning without the need for task-specific hyperparameter tuning.
\begin{figure*}[!t]
    \centering
    \includegraphics[width=\linewidth]{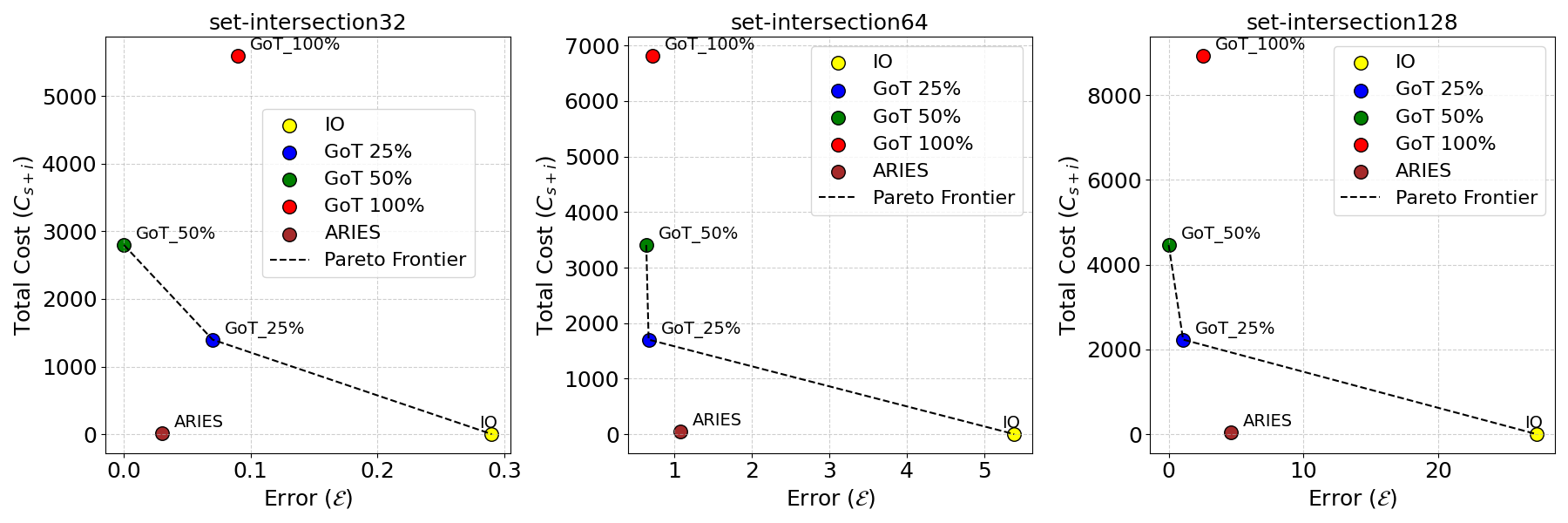}
    \caption{Pareto frontiers in total query cost ($C_{s+i}$) and task error ($\mathcal{E}$) for set intersection tasks at various difficulty levels. The total cost is the number of queries expended at search and inference time. Llama-3.1-405B was used for the reasoning and policy agents. Our results (ARIES) have pushed the Pareto frontiers forward in each task.}
    \label{fig:sets-pareto}
\end{figure*}

\subsubsection{Set Intersection}

In Figure \ref{fig:sets-pareto}, we plot a Pareto curve showing viable trade-off points in task error and query cost for the set intersection task. Our approach extends the existing Pareto frontier constructed by considering static schedule baselines and direct prompting. In the set-intersection32 task, we achieve a $2.3\times$ error reduction relative to $\text{GoT}_{25}$ while also achieving $116\times$ lower overall cost.

\subsection{Transition Probability Profiling} \label{section:transition-probabilities}

In this section, we estimate the transition probabilities for each thought graph transformation across a number of tasks to gain insight into factors impacting a thought graph formulation of each reasoning problem. For $\phi_{ref}$, we define a successful transition when $\mathcal{E} = 0$ for the resulting node, considering only cases when the transformation is executed on nodes previously containing errors. In transformations requiring LLM calls, the transition probability between two states is a random process governed by the token distribution parametrized by the LLM. When LLM calls are not required, i.e. the transformation is implemented through simple node manipulation, the transition probability is 1.

\begin{table*}[t]
    \centering
    \caption{Failure mode 1 results. Mean value of the error $\mathcal{E}$ ($\downarrow$) for benchmarks with low decomposition depth. Llama-3.1-70B was used for the reasoning and policy agents.}

        \begin{tabular}{c|ccccc}
        \toprule
        \textbf{Method} 
        & Direct Prompting
        & $\text{GoT}_{25\%}$
        & $\text{GoT}_{50\%}$
        & $\text{GoT}_{100\%}$
        & ARIES
        \\
        \midrule
        sorting32 
        & 2.2
        & 0.82 
        & 0.95 
        & \textbf{0.73} 
        & 1.29 \\
        % sorting64 
        % & \textbf{2.22} & 2.74 & 3.46 & 9.15 & 10.04 \\
        % sorting128 
        % & 13.96 & \textbf{12.65} & 18.65 & 32.74 & 31.79 \\
        set-intersection32  
        & 1.05
        & 0.41 & \textbf{0.0} & 0.37 & 1.22
        \\
        % set-intersection64 
        % & 0.67 & \textbf{0.64} & 0.72 & 1.08 & 7.34 \\
        % set-intersection128 
        % & \textbf{1.07} & 0 & 2.54 & 4.62 & 22.97 \\
        \bottomrule
        \end{tabular}
        \label{table:failure_size}

\end{table*}
\begin{table*}[t]
    \centering
    \caption{Failure mode 2 results. Mean value of the error $\mathcal{E}$ ($\downarrow$) and search cost $C$ in terms of number of queries ($\downarrow$). Both the reasoning and policy agents are LLaMA-405B.}

        % \resizebox{\textwidth}{!}{ % This scales the table
            \begin{tabular}{c|cc|cc|cc|cc|cc}
            \toprule
            \textbf{Method} 
            & \multicolumn{2}{c|}{Direct Prompting}
            & \multicolumn{2}{c|}{$\text{GoT}_{25\%}$}
            & \multicolumn{2}{c|}{$\text{GoT}_{50\%}$}
            & \multicolumn{2}{c|}{$\text{GoT}_{100\%}$}
            & \multicolumn{2}{c}{ARIES}
            \\
            Metrics
            &$\mathcal{E}$ & $C$
            &$\mathcal{E}$ & $C$
            &$\mathcal{E}$ & $C$
            &$\mathcal{E}$ & $C$
            &$\mathcal{E}$ & $C$
            \\
            \midrule
            sorting32 
            & 0.6
            & \textbf{1}
            & 0.74 
            & 825
            & 0.82 
            & 1650
            & 0.28 
            & 3300
            & \textbf{0.22}
            & 20 \\
            sorting64 
            & 5.07
            & \textbf{1}
            & 2.22
            & 1671
            & \textbf{2.74}
            & 3343
            & 3.46
            & 6687
            & 9.15
            & 48 \\
            sorting128 
            & 12.75
            & \textbf{1}
            & 13.96
            & 2444
            & \textbf{12.65}
            & 4888
            & 18.65
            & 9776
            & 32.74
            & 48 \\
            % set-intersection64 
            % & 5.38
            % & \textbf{1}
            % & 0.67
            % & 1703
            % & \textbf{0.64}
            % & 3407
            % & 0.72
            % & 6815
            % & 1.08
            % & 48 \\
            % set-intersection128 
            % & 27.29
            % & \textbf{1}
            % & 1.07
            % & 2231
            % & \textbf{0.0}
            % & 4463
            % & 2.54
            % & 8927
            % & 4.62
            % & 48 \\
            \bottomrule
            \end{tabular}
        % }
        \label{table:sorting-results}

\end{table*}

\begin{table}[h]
    \centering
    \caption{Esimated transition probabilities for each thought graph transformation, taken as the number of successful state transitions in a static schedule.}
    \label{table:transition-probabilities}

    \begin{tabular}{ccccc}
    \toprule
     & $\mathbf{\phi_{sol}}$ & $\mathbf{\phi_{ref}}$ & $\mathbf{\phi_{red}}$ & $\mathbf{\phi_{agg}}$ \\ 
    \midrule
    \textbf{HumanEval} & 0.77 & 0.29 & 1 & 1 \\
    \textbf{sorting32} & 0.57 & 0.12 & 1 & 0.60 \\
    \textbf{set-intersection32} & 0.75 & 0.71 & 1 & 1 \\
    \bottomrule
    \end{tabular}
\end{table}

The results are summarized in \Cref{table:transition-probabilities}. We observe the refinement transformation has notably low success probability, particularly in coding and sorting tasks. Additionally, sorting is the only task with non-deterministic aggregation, which is a potential error source. We note that the performance of a thought graph formulation depends on the ability of the policy agent to capture the success profile of various transformations for a task, and adapt the exploration strategy accordingly.

\subsection{Failure Modes} \label{section:failure-modes}

In this section, we perform a number of empirical studies aiming to understand the main limiting factors impacting the performance of LLM policy agents on interactive thought graphs. We find there are two major failure modes, described as follows.

\noindent
\textit{\myfancylabel~~Failure mode 1: LLM Parameter Count}

\noindent
We find that LLMs with insufficiently large parameter sizes exhibit limited performance when utilized as policy agents on thought graph environments. We deploy Llama-3.1-70B as policy and reasoning agents in sorting and set intersection tasks, against which the larger LLM (Llama-405B) was shown to perform well as a policy agent. As shown in \Cref{table:failure_size}, LLM-guided graph exploration (ARIES) did not outperform static schedule baselines in this scenario. These findings are consistent with \citep{emergent_abilities}, which demonstrated that zero-shot chain-of-thought reasoning abilities emerges in models beyond 175B parameters.

% Interestingly, we observe that further search trials at times cause an increase in error values, i.e. the condition $\text{GoT}_{\text{100}}~<~\text{GoT}_{\text{50}}~<~\text{GoT}_{\text{25}}$ is not met. This is understood due to the multi-objective search formulation, causing the optimizer to prioritize configurations with improved query efficiency, despite incurring an error cost.
\noindent
\textit{\myfancylabel~~Failure mode 2: Decomposition Depth}

\noindent
We examine the impact of decomposition depth by analyzing the results in the sorting task, shown in \Cref{table:sorting-results}. We observe LLM policy agents lead to a $21\%$ performance improvement relative to the most optimized static baseline in sorting32, which has a decomposition depth of 2. However, as discussed in \Cref{section:transition-probabilities}, the sorting task presents a particular challenge due to the lower success probability of the aggregation transformation. As the complexity and decomposition depth of a task increases, the policy agent is required to apply a higher number of aggregation transformations. Therefore, we observe up to $4.12\times$ and $2.6\times$ performance deterioration in sorting64 and sorting128, respectively. Through empirical analysis, we observe that in the latter tasks, the $\phi_{agg}$ transformation constitutes $86\%$ and $68\%$ of all policy agent errors, respectively. As such, we conclude that high decomposition depths present a significant failure mode for LLM-guided thought graph exploration, particularly in tasks with low success transition probabilities for the aggregation transformation.

\section{Ablation Studies} \label{section:ablation}
% Ablation study 
% 1. With CoT, the impact of ensemble size is more significant. There is an obvious trend in reduction in mean error when using more LLMs agents to decide in actions. 
% a. The effectivness of ensemble size is related to the difficulty of the problem
% b. However, the performance enhancement from increasing ensemble size diminished after a certain threshold.

% 2. Without CoT, the impact of ensemble size is inconclusive. Sometimes an increase in ensemble size will even cause an increase in error.

\begin{figure}[!t]
    \centering
    \includegraphics[width=0.5\linewidth]{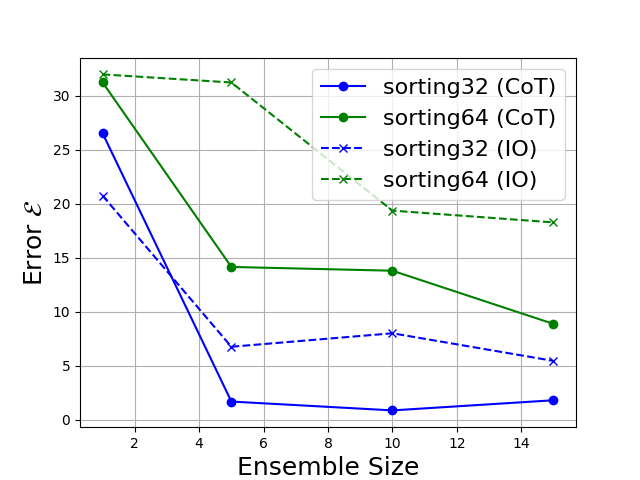}
    \caption{Mean error (y-axis) obtained in the sorting32 task over a sweep of ensemble sizes (x-axis). Llama-3.1-70B was used as the policy agent.}
    \label{fig:ablation-study}
\end{figure}

As discussed in Section \ref{section:autonomous-graph-exploration}, two factors that impact the performance of LLMs as policy agents in interactive thought graph environments are the size of the ensemble and the use of chain of thought reasoning to enhance the planning abilities of the policy agent. In this section, we aim to understand the impact of each factor by evaluating sorting tasks over a range of ensemble sizes from 1 to 15, with and without CoT prompting in the policy agent.

% As shown in \Cref{fig:ablation-study}, we see a large performance improvement by scaling up the ensemble size up to 5, with diminishing returns thereafter. However, the trend is less clear when CoT prompting is not integrated, with higher ensemble sizes occasionally yielding worse performance. As such, we show the necessity of CoT prompting to enhance the ability of the LLM policy agent to adapt from feedback and drive thought graph transformations.

As shown in \Cref{fig:ablation-study}, as the ensemble size increases to 5, CoT prompting leads to large performance improvements, though the benefits start diminishing beyond this point. Without CoT prompting, the trend is less consistent, and larger ensemble sizes sometimes yield worse performance. Additionally, errors without CoT are higher for both tasks at any ensemble size. This highlights the necessity of CoT prompting in enhancing the LLM policy agent’s ability to adapt from feedback and drive thought graph transformations.

\section{Conclusion}

We introduce ARIES, a multi-agent architecture for topological reasoning. By viewing thought graph transformations as actions in a Markov decision process, we show off-the-shelf LLMs can drive efficient action policies without task-specific tuning. We show up to $29\%$ higher accuracy on HumanEval while reducing inference costs by $35\%$ compared to static schedules. We identified two key limitations: insufficient model size and excessive decomposition depth on the task at hand. These constraints indicate that while LLMs show promise as reasoning agents, their effectiveness depends on parameter count and task complexity.

\newpage

\bibliography{main}
\bibliographystyle{conference}

\newpage

\appendix

\section{Limitations}

\subsection{Assumptions and Robustness}

The ARIES framework introduces a novel approach to reasoning with large language models (LLMs) through interactive thought graph environments. However, several strong assumptions underlie our methodology. Firstly, we assume that thought graph transformations can be effectively modeled as a Markov decision process (MDP) with well-defined state transitions. While this formulation enables structured reasoning, it may not fully capture the complexities of more ambiguous or highly interconnected problems. Additionally, our approach assumes that off-the-shelf LLMs can act as reliable policy agents without additional fine-tuning. This assumption holds for certain problem domains but may degrade in tasks requiring domain-specific knowledge or long-horizon planning.

Our empirical evaluation is constrained to specific reasoning tasks, including HumanEval, list sorting, and set intersection. While these benchmarks serve as valuable test cases for structured reasoning, they do not necessarily generalize to all problem types, particularly those with weakly defined intermediate states or multi-modal reasoning requirements. Furthermore, our evaluation primarily focuses on LLaMA-3.1 models, and results may not be directly transferable to other architectures.

\subsection{Potential Risks}

The ARIES framework introduces both opportunities and challenges in autonomous reasoning. One primary risk is the potential for incorrect or biased reasoning paths due to the stochastic nature of LLM-generated decisions. Although our policy agent ensembles mitigate some of this variability, they do not fully eliminate erroneous transformations, particularly in deeper decomposition settings. The framework’s reliance on existing LLMs also means that any biases present in the underlying models could propagate into the reasoning process, potentially leading to unfair or misleading outcomes.

Another concern is the environmental impact associated with inference-heavy approaches. While ARIES improves query efficiency relative to static transformation schedules, it still necessitates a significant number of LLM queries to achieve high accuracy. As LLMs scale, the energy consumption required for these inference tasks could become a sustainability concern, particularly in high-throughput applications.

\subsection{Failure Modes}

Our empirical findings highlight two major failure modes: (1) inadequate LLM parameter sizes and (2) increasing decomposition depth. Smaller models (e.g., LLaMA-3.1-70B) struggle to act as policy agents effectively, demonstrating subpar reasoning capabilities compared to larger counterparts. This suggests that autonomous policy-driven thought graph exploration may require models beyond a certain scale threshold to function reliably. Additionally, as the depth of problem decomposition increases, ARIES exhibits a decline in performance, primarily due to errors in aggregating intermediate solutions. This limitation indicates that current LLMs may have difficulties managing extended reasoning chains, which presents a barrier to scalability.

\section{Thought Graph Transformations} \label{appendix:thought_graph_transformations}

\begin{table*}[t]
    \centering
    \caption{Thought graph transformations. Each transformation is defined as $\phi (G_\tau, m, S) = (V \cup V^+ \setminus V^-, E \cup E^+ \setminus E^-)$, where $G_\tau = (V, E)$ is a thought graph, $S \subset V$ is a subset of nodes, $m$ is the multiplicity (number of attempts), and $\mathcal{E}$, $\mathcal{R}$, $\mathcal{A}$ represent arbitrary functions for node expansion, refinement and aggregation, respectively. The sets $V^+, V^-, E^+, E^-$ are defined as follows.}
    \label{table:transformations}
    \vspace{0.5em}
    \resizebox{\textwidth}{!}{ % This scales the table
    
        \begin{tabular}{lccccc}
        \toprule
        \textbf{Transformation} & \textbf{Symbol} & $\mathbf{V^+}$ & $\mathbf{V^-}$ & $\mathbf{E^+}$ & $\mathbf{E^-}$ \\ 
        \midrule
        
        {Decompose} & $\phi_{dec}$ & $\{\mathcal{E}(v) | v \in S\}$ & $\emptyset$ & $\{ (u, v) | u \in S, v \in V^+ \}$ & $\emptyset$ \\
        
        {Solve} & $\phi_{sol}$ & $\{\mathcal{S}(v) | v \in S\}$ & $\emptyset$ & $\{ (u, v) | u \in S, v \in V^+ \}$ & $\emptyset$ \\
    
        {Refine} & $\phi_{ref}$ & $\{\mathcal{R}(t) | t \in S\}$ & $\emptyset$ & $\{ (u, v) | u \in S, v \in V^+ \}$ & $\emptyset$ \\
        %  \midrule
        
        {Reduce} & $\phi_{red}$ & $\emptyset$ & $S$ & $\emptyset$ & $\{ (u, v) | u \in S \vee v \in S \}$ \\
        % \midrule
        
        {Aggregate} & $\phi_{agg}$ & $\mathcal{A}(S)$ & $\emptyset$ & $\{ (u, v) | u \in S, v \in V^+ \}$ & $\emptyset$ \\
        \bottomrule
        \end{tabular}
    }
\end{table*}

The full set of considered transformations is shown in Table \ref{table:transformations}.

\section{Static Schedule Parameter Search} \label{section:search}

As described in Section \ref{section:topological-reasoning}, a static transformation can be characterized using a set of discrete parameters. We ran bayesian search using using Tree-structured Parzen Estimator (TPE) sampling to determine each parameter, establishing strong baselines for each task.

\begin{table} [h!]
    \centering
    \caption{Search space for each parameter characterizing a static transformation.}
    \label{table:got-search-space}
    
    \begin{tabular}{ccc}
    \toprule
    \textbf{} & & \textbf{Search}  \\ 
    \textbf{} & \textbf{Parameter} & \textbf{Space} \\ 
    \midrule
    $R_{ed}$ & Allow reduction & $\{0, 1\}$ \\
    $R_{ef}$ & Allow refinement & $\{0, 1\}$ \\
    $S^m$ & Solve multiplicity & $\{1, 5, 10, 15, 20\}$ \\
    $A^m$ & Aggregate multiplicity & $\{1, 5, 10, 15, 20\}$ \\
    $R_{ef}^m$ & Refine multiplicity & $\{1, 5, 10, 15, 20\}$ \\
    \bottomrule
    \end{tabular}
\end{table}

The search space is shown in Table \ref{table:got-search-space}. We run multi-objective search to concurrently minimize the task-specific error function $\mathcal{E}$ (Section \ref{section:benchmarks}) and associated cost, measured as $|\Phi (\omega)|$ where $\Phi (\omega) = (\phi_0, \dots, \phi_m)$ is a tuple enumerating thought graph transformations, as a function of the schedule parameters $\omega \in \Omega$, where $\Omega$ is the search space. Note that $|\Phi (\omega)|$ correlates with the number of LLM queries, meaning this formulation aims to minimize exploration cost.

In selecting parameter configurations, we use the cost function in Equation \ref{eq:cost-function}, such that the objectives of cost and error minimization are balanced through the scalar constant $\alpha \in (0, 1)$. We aim to assign equal importance to the cost and error objectives by tuning $\alpha$ independently for each task such that the mean value of the first term matches the second term, i.e. $\alpha E \left [ \mathcal{E} \right ] = (1-\alpha) E \left [ |\Phi (\omega)|) \right ] $, or equivalently $\alpha = \frac{E[|\Phi (\omega)|]}{E[\mathcal{E} + |\Phi (\omega)|]}$ where $E$ denotes the expected value. The expectations are obtained with random sampling.
\begin{equation} \label{eq:cost-function}
    \underset{\omega}{\min} \left [ \alpha \mathcal{E} + (1 - \alpha) |\Phi (\omega)| \right ]
\end{equation}

Search was conducted separately on Llama-3.1-70B and Llama-3.1-405B. For sorting and set intersection tasks, search is conducted separately for each difficulty level, ensuring the chosen parameters are adapted to the task. Note that we present three search checkpoints $\text{GoT}_{n}$ for $n \in \{25, 50, 100\}$, where $n$ corresponds to the percentage of trials until convergence. We define the convergeance point as the first iteration where a rolling window $J$ of size 20 matches the condition $J^k = J^{k-1}$. This enables comparing our proposed LLM-guided approach to optimized search schedules at various search budgets.

\begin{table}[h!]
    \centering
    \caption{Results from GoT static schedule parameter search on Llama-3.1-405B.}
    \label{table:got-search-results}

    \begin{tabular}{ccccc}
    \toprule
    \textbf{Task} & \textbf{Alpha ($\alpha$)} & $\textbf{GoT}_{\textbf{25}}$ & $\textbf{GoT}_{\textbf{50}}$ & $\textbf{GoT}_{\textbf{100}}$ \\ 
    \midrule
    sorting32 & 0.99 & 0.38 & 0.38 & 0.37 \\
    sorting64 & 0.96 & 4.85 & 4.49 & 3.84 \\
    sorting128 & 0.84 & 28.76 & 25.76 & 24.36 \\
    set32 & 0.99 & 0.16 & 0.16 & 0.12 \\
    set64 & 0.99 & 0.71 & 0.51 & 0.31 \\
    set128 & 0.98 & 3.51 & 3.51 & 2.99 \\
    % keyword & 0.99 & 0.96 & 0.96 & 0.32 \\
    \bottomrule
    \end{tabular}
\end{table}

% \begin{figure*}[b]
%     \label{fig:query-efficiency}
%     \centering
%     \caption{Growth plot for set-intersection32 and sorting32 task accuracy ($\uparrow$) with the number of reasoning agent LLM queries. Llama-3.1-405B was used for both reasoning and planning agents across all GoT baselines and the LLM-guided approach. Note that policy agent queries are not considered, i.e. the queries required to analyze the thought graph and sample the action space.}
%     \includegraphics[width=\linewidth]{imgs/query-efficiency.pdf}
% \end{figure*}

% , while results for Llama-3.1-70B are shown in Appendix \ref{section:llama-70B-search-results}
The complete search results for Llama-3.1-405B are shown in Table \ref{table:got-search-results}. It can be seen that tasks with higher decomposition depth incur lower values of $\alpha$ due to the higher magnitude of the error function. sorting64, sorting128 and set-intersection64 show a smooth decline in the cost function, while the remaining tasks remain at local minima until close to the end of the search. The non-convexity of the search space highlights the cost associated to optimize the parameter set associated with static transformations.

\begin{table*}
    \label{table:results}
    \centering
    \caption{Core results for topological reasoning across all tasks and models. We show the mean value of the score function $\mathcal{E}$ ($\downarrow$), which is defined for each task in Section \ref{section:experiments}. $\text{GoT}_{100}$, $\text{GoT}_{50}$, $\text{GoT}_{25}$ represent the obtained values from static schedule parameters obtained at convergeance, 50\% and 25\% of convergeance trials, respectively.}

    \resizebox{\textwidth}{!}{ % This scales the table
    
        \begin{tabular}{c|cccc|cccc}
        \toprule
        \textbf{Task} & \multicolumn{4}{c|}{\textbf{Llama-70b}} & \multicolumn{4}{c}{\textbf{Llama-405b}}  \\
        \midrule
        & $\text{GoT}_{25}$ & $\text{GoT}_{50}$ & $\text{GoT}_{100}$ & $\text{GoT}_{\text{LLM}}$ & $\text{GoT}_{25}$ & $\text{GoT}_{50}$ & $\text{GoT}_{100}$ & $\text{GoT}_{\text{LLM}}$ \\
        \midrule
        sorting32 & 0.82 & 0.95 & \textbf{0.73} & 1.29 & 0.74 & 0.82 & 0.28 & \textbf{0.22} \\
        sorting64 & 4.73 & 4.73 & \textbf{4.64} & 10.04 & \textbf{2.22} & 2.74 & 3.46 & 9.15 \\
        sorting128 & 16.18 & \textbf{13.86} & 16.07 & 31.79 & 13.96 & \textbf{12.65} & 18.65 & 32.74 \\
        set-intersection32 & 0.41 & \textbf{0.0} & 0.37 & 1.22 & 0.07 & 0.0 & 0.09 & \textbf{0.03} \\
        set-intersection64 & 3.40 & 2.66 & \textbf{1.27} & 7.34 & 0.67 & \textbf{0.64} & 0.72 & 1.08 \\
        set-intersection128 & 13.23 & 12.92 & \textbf{12.73} & 22.98 & \textbf{1.07} & 0 & 2.54 & 4.62 \\
        % keyword-counting & 0.05 & 0.14 & 0.15 & 0 & 0.05 & 0.05 & 0.04 & 0 \\
        \bottomrule
        \end{tabular}
    }
\end{table*}

% \newpage
\section{Benchmarks} \label{section:benchmarks}

We choose two popular tasks for topological reasoning with LLMs, which are amenable to a divide-and-conquer strategy (i.e. decomposition, solving subproblems and merging): list sorting and set intersection. Despite their simplicity, prior works have shown that these tasks are extremely challenging for LLMs with direct prompting \citep{got}. 
% \az{If space is a problem, you can move the rest of this section into appendix.}

\textbf{Sorting}: involves sorting a list of numbers between $0$ and $9$ in ascending order. The error function $\mathcal{E} = X + Y$ has its subterms defined in Equation~\ref{eq:sorting}, where $a$ is the input list and $b$ is a candidate solution. $X$ corresponds to the number of incorrectly sorted pairs, while $Y$ corresponds to the frequency difference between $a$ and $b$ for each digit. 
\begin{equation}
    \begin{aligned}
        X &= \sum_{i=1}^{m-1} \text{sign}(\max(b_i - b_{i+1}, 0)) \\
        Y &= \sum_{i=0}^{9} \left| \left| \{b_p : b_p = i\} \right| - \left| \{a_q : a_q = i\} \right| \right|
    \end{aligned}
    \label{eq:sorting}
\end{equation}

\textbf{Set Intersection}: involves finding the intersection of sets $A$ and $B$. The error function is defined in Equation~\ref{eq:set-intersection}, where $C$ is the candidate solution. The first and second terms correspond to missing and extra elements, respectively.
\begin{equation} \label{eq:set-intersection}
    \mathcal{E} = |(A \cap B) \setminus C| + |C \setminus (A \cap B)|
\end{equation}

\end{document}